%% file: main.tex
\documentclass[10pt,twocolumn,letterpaper]{article}

\usepackage{cvpr}              

\usepackage{multirow}
\usepackage{array}
\usepackage{enumitem}
\usepackage{makecell}
\usepackage{tabularx}
\usepackage{caption}

\definecolor{cvprblue}{rgb}{0.21,0.49,0.74}
\usepackage[pagebackref,breaklinks,colorlinks,allcolors=cvprblue]{hyperref}

\def\paperID{14957} 
\def\confName{CVPR}
\def\confYear{2026}

\title{AutoCut: End-to-end advertisement video editing based on multimodal discretization and controllable generation}

\author{
Milton Zhou$^{1,2,}$\thanks{Equal contribution. Work done during an internship at Kuaishou.} \quad
Sizhong Qin$^{1,2,}$\footnotemark[1] \quad
Yongzhi Li$^{2, }$\thanks{Project Leader, $^\ddagger$Corresponding author.} \quad
Quan Chen$^{2, \ddagger}$ \quad
Peng Jiang$^{2}$ \\
$^1$Tsinghua University \quad $^2$Kuaishou Technology \\
{\tt\small \{zhoukx23, qsz23\}@mails.tsinghua.edu.cn} \\
{\tt\small \{liyongzhi03,chenquan06,jiangpeng\}@kuaishou.com}
}

\begin{document}

\maketitle

\input{sec/0_abstract}    
\input{sec/1_intro}
\input{sec/2_related_work}
\input{sec/3_method}
\input{sec/4_exp}

\input{sec/5_discussion}
\input{sec/6_conclusion}

{
    \small
    \bibliographystyle{ieeenat_fullname}
    \bibliography{main}
}

\input{sec/7_suppl}

\end{document}

%% file: sec/0_abstract.tex
\begin{abstract}
Short-form videos have become a primary medium for digital advertising, requiring scalable and efficient content creation.
However, current workflows and AI tools remain disjoint and modality-specific, leading to high production costs and low overall efficiency. 
To address this issue, we propose \textbf{AutoCut}, an end-to-end advertisement video editing framework based on multimodal discretization and controllable editing. AutoCut employs dedicated encoders to extract video and audio features, then applies residual vector quantization to discretize them into unified tokens aligned with textual representations, constructing a shared video–audio–text token space. Built upon a foundation model, we further develop a multimodal large language model for video editing through combined multimodal alignment and supervised fine-tuning, supporting tasks covering video selection and ordering, script generation, and background music selection within a unified editing framework. 
Finally, a complete production pipeline converts the predicted token sequences into deployable long video outputs.
Experiments on real-world advertisement datasets show that AutoCut reduces production cost and iteration time while substantially improving consistency and controllability, paving the way for scalable video creation.
Code and data are available at: \url{https://github.com/AdAutoCut/Autocut}
\end{abstract}

%% file: sec/1_intro.tex
\section{Introduction}
\label{sec:intro}

\begin{figure}[t]
  \centering
  \includegraphics[width=\linewidth]{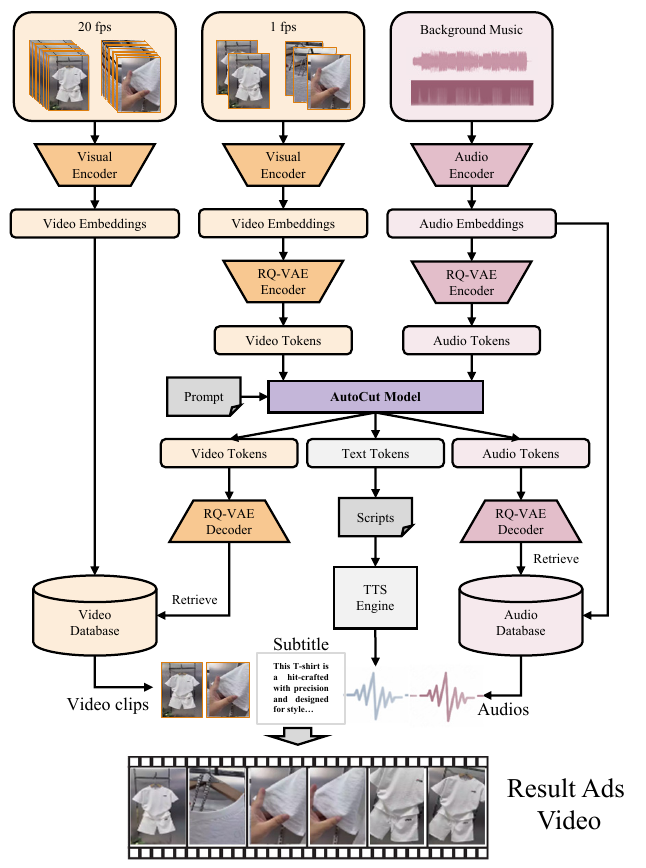}
  \caption{
\textbf{Overview of the AutoCut framework.}
Low–fps frames are tokenized for efficient multimodal reasoning, while high–fps frames are kept for accurate visual matching and clip retrieval. 
Given optional inputs, AutoCut predicts video, text, and audio tokens, which are decoded or retrieved to compose the final advertisement video.
}
  \label{fig:Teaser}
\vspace{-0.2in}
\end{figure}

Advertising videos have become one of the most influential forms of digital marketing, allowing brands to communicate persuasive narratives through multimodal visual storytelling. However, producing short-form advertisement videos remains a costly and labor-intensive process that involves scripting, material shooting, editing, and post-production refinement. This multi-stage workflow requires professional expertise and substantial manual effort, creating high entry barriers for small enterprises and non-expert creators~\cite{weiIntelligentAdvertisementShort2022,zhangVideoDiscoveryAutomaticShortVideo2021}.  

To alleviate these challenges, recent studies have explored intelligent systems that automate parts of the video creation pipeline. Early works such as \textit{VideoDiscovery}~\cite{zhangVideoDiscoveryAutomaticShortVideo2021} and Wei \textit{et al.}~\cite{weiIntelligentAdvertisementShort2022} proposed multimodal retrieval frameworks that generate advertising videos directly from descriptive copy, reducing manual effort but still relying on rigid, predefined templates. Lin \textit{et al.}~\cite{linAutomatedMultiModalVideo2021} and Tang \textit{et al.}~\cite{tangMultimodalSegmentAssemblage2023} further introduced multimodal editing and segment-assemblage networks to improve visual coherence. Nevertheless, these modular approaches remain constrained by handcrafted rules and offer limited global controllability.  

With the emergence of multimodal large language models (MLLMs), new opportunities have arisen for end-to-end video understanding and editing.
Cheng \textit{et al.}~\cite{chengTexttoEditControllableEndtoEnd2025} proposed a text-to-edit paradigm that enables users to create videos through natural-language instructions, while Qian \textit{et al.}~\cite{qianVCLLMAutomatedAdvertisement2025} introduced VC-LLM, which achieves human-comparable advertisement video generation via multi-resolution spatial–temporal reasoning. These studies highlight the potential of MLLMs to unify perception, understanding, and creation within a single framework.  However, directly applying standard MLLM models to advertising creation scenarios is limited by constrained context window sizes, making it challenging to handle large-scale video retrieval and editing tasks.

Nevertheless, existing systems still face three major obstacles that hinder scalable, controllable, and coherent video creation:  
\begin{itemize}
    \item \textbf{Loose multimodal coupling.} Representations across modalities are often weakly aligned, preventing unified reasoning among video, audio, and textual signals.  
    \item \textbf{Lack of interpretable control.} Most models do not provide structured or discrete representations, making it difficult to adjust narrative rhythm, temporal composition, and content emphasis in a controllable manner.  
    \item \textbf{Fragmented understanding and generation.} Current pipelines treat multimodal understanding and generation as separate processes, leading to inconsistent optimization and unstable editing quality.  
\end{itemize}

In this work, we present AutoCut, an end-to-end intelligent advertisement video editing framework that integrates \emph{multimodal discretization} and \emph{controllable editing}.
AutoCut unifies video, audio, and textual information within a shared discrete token space, enabling fine-grained multimodal reasoning and precise editing control. By leveraging large-scale real-world advertisement data, AutoCut significantly reduces production costs while enhancing coherence, controllability, and creative quality.  

Our main contributions are summarized as follows:  
\begin{itemize}
    \item Proposes the first unified framework that bridges multimodal understanding and controllable editing for advertisement video creation.
    \item Designs a multimodal discretization strategy that transforms video, audio, and text signals into interpretable tokens, enabling controllable and semantically aligned video editing.  
    \item Establishes a large-scale advertisement video dataset and demonstrates through extensive experiments that AutoCut achieves superior coherence, editing consistency, and content controllability compared with existing methods.  
\end{itemize}

%% file: sec/2_related_work.tex
\section{Related Work}
\label{sec:related_work}

\begin{figure*}[t]
  \centering
  \includegraphics[width=\linewidth]{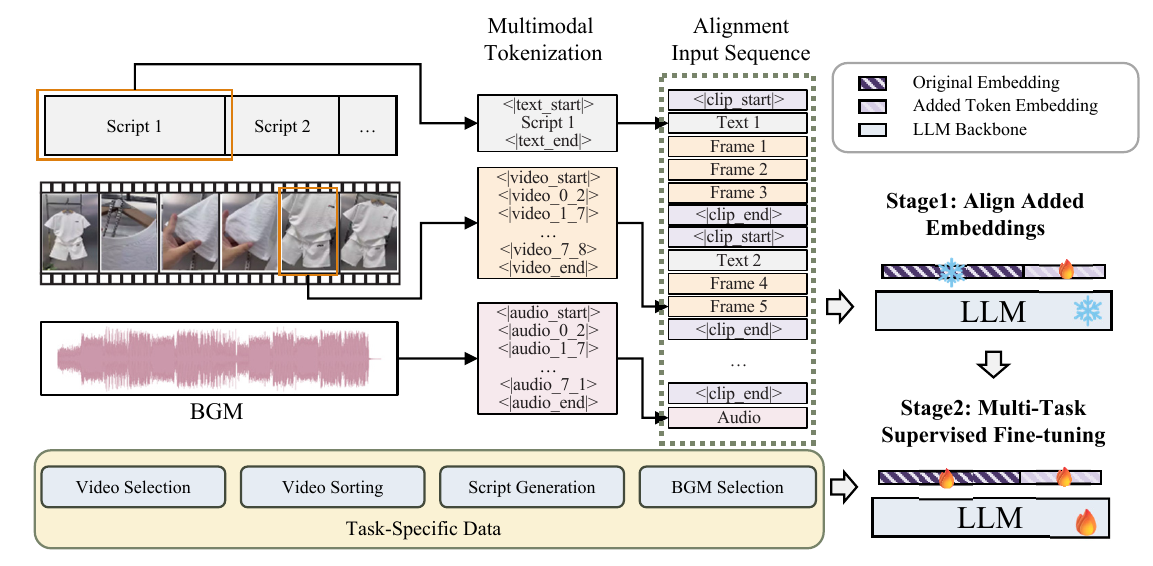}
  \caption{
\textbf{Overview of the proposed AutoCut framework.}
Multimodal tokenization converts scripts, frames, and audio into unified discrete tokens, which are organized into an alignment input sequence. 
Stage~1 performs multimodal alignment on large-scale data to align the added token embeddings with the LLM backbone. 
Stage~2 applies task-specific SFT for video selection, video sorting, script generation, and BGM selection.
}
  \label{fig:Method}
\vspace{-0.2in}
\end{figure*}

\subsection{Template-Based Methods}
Early research formalized video editing as rule-based or template-based composition to maintain narrative coherence and stylistic consistency.  
Ahanger and Little~\cite{ahangerAutomaticCompositionTechniques1998} introduced metadata-driven dynamic composition, while systems such as \textit{QuickCut}~\cite{truongQuickCutInteractiveTool2016} and multi-camera frameworks~\cite{arevAutomaticEditingFootage2014} extended these ideas to dialogue-driven and viewpoint-driven assembly.  
Commercial tools including Magisto, CapCut, and iMovie encoded cinematic conventions into templates that automatically determine shot order, transition types, and pacing.  
\textit{Lu et al.}~\cite{luComputationalProductPresentation2020} added rule-constrained optimization for product-oriented video editing.  
Although these systems maintain structure, their dependence on predefined rules restricts flexibility.  
Recent datasets such as \textit{AVE}~\cite{argawAnatomyVideoEditing2022} and shot-order benchmarks~\cite{liShotSequenceOrdering2025} have enabled data-driven extensions beyond handcrafted templates.

\vspace{3pt}
\subsection{Retrieval-Based Methods}
Retrieval-based approaches frame editing as multimodal search followed by clip assembly.  
Systems such as \textit{VideoDiscovery}~\cite{zhangVideoDiscoveryAutomaticShortVideo2021} and \textit{Wei et al.}~\cite{weiIntelligentAdvertisementShort2022} retrieve footage aligned with textual descriptions for advertisement creation.  
Script-driven methods including \textit{Write-A-Video}~\cite{wangWriteavideoComputationalVideo2019}, \textit{B-Script}~\cite{huberBScriptTranscriptbasedBroll2019}, and \textit{Transcript-to-Video}~\cite{xiongTranscriptVideoEfficient2022} improve textual-visual alignment, while \textit{Story-Driven Editing}~\cite{wangStorydrivenVideoEditing2021} enhances temporal organization.  
For multi-source workflows, \textit{DeepQAMVS}~\cite{messaoudDeepQAMVSQueryAwareHierarchical2021} and \textit{Condensed Movies}~\cite{bainCondensedMoviesStory2020} address summarization and narrative reconstruction.  
These methods provide efficient clip-level editing but remain limited by dataset coverage and retrieval accuracy.  
\textit{VEU-Bench}~\cite{liVEUBenchComprehensiveUnderstanding2025} offers complementary evaluation for understanding editing operations and narrative structure.

\vspace{3pt}
\subsection{Generative Methods}
Generative approaches synthesize or modify content beyond direct retrieval.  
Early research explored unsupervised highlight detection~\cite{yangUnsupervisedExtractionVideo2015,xiongLessMoreLearning2019,rochanAdaptiveVideoHighlight2020} and multimodal summarization~\cite{zhaoHierarchicalMultimodalTransformer2022,merlerAutomaticCurationSports2019}.  
M-SAN~\cite{tangMultimodalSegmentAssemblage2023} introduced importance and coherence rewards for advertisement editing, and multimodal encoders~\cite{guoMultimodalRepresentationLearning2021} improved cross-modal representation learning.  
Interactive systems such as \textit{ReelFramer}~\cite{wangReelFramerHumanAICoCreation2024}, \textit{ChunkyEdit}~\cite{leakeChunkyEditTextfirstVideo2024}, and \textit{ExpressEdit}~\cite{tilekbayExpressEditVideoEditing2024} support text-guided and sketch-guided content manipulation.  
Hybrid pipelines such as \textit{Computational Video Editing}~\cite{leakeComputationalVideoEditing2017} integrate retrieval and generation using semantic alignment and HMM-based optimization.  
Benchmarks including \textit{Shot2Story}~\cite{hanShot2StoryNewBenchmark2025} and \textit{Synchronized Video Storytelling}~\cite{yangSynchronizedVideoStorytelling2024} further evaluate story-level controllability.  
Despite these advances, generative approaches still struggle with temporal consistency, realism, and multi-scene stability.

\subsection{MLLM-Based Methods}
\vspace{-0.05in}
Multimodal Large Language Models shift video editing from pattern-driven workflows toward reasoning-driven orchestration across modalities.  
Systems such as \textit{LAVE}~\cite{wangLAVELLMPoweredAgent2024}, \textit{Ding et al.}~\cite{dingPromptDrivenAgenticVideo2025}, and \textit{LITA}~\cite{huangLITALanguageInstructed2024} incorporate temporal cues such as time tokens or SlowFast~\cite{feichtenhoferSlowFastNetworksVideo2019} features for localized editing.  
Subsequent methods including \textit{Text-to-Edit}~\cite{chengTexttoEditControllableEndtoEnd2025}, \textit{VC-LLM}~\cite{qianVCLLMAutomatedAdvertisement2025}, and \textit{Tree-of-AdEditor}~\cite{zhangTreeofAdEditorHeuristicTree2025} apply high-level reasoning to multi-scene advertisement assembly.  
\textit{Shots2Stories}~\cite{liShotsStoriesLLMAssisted2025} further investigates LLM-guided shot ordering and clip-level coherence.

A related direction combines MLLMs with diffusion models for pixel-level manipulation.  
Systems such as \textit{InstructX}~\cite{mouInstructXUnifiedVisual2025}, \textit{UNIC}~\cite{yeUNICUnifiedInContext2025}, \textit{UniVideo}~\cite{weiUniVideoUnifiedUnderstanding2025}, and \textit{VEGGIE}~\cite{yuVEGGIEInstructionalEditing2025} focus on instruction-driven visual transformation rather than clip selection, sequencing, or narrative assembly.

Although these methods demonstrate strong multimodal reasoning, most rely on separate modules for understanding, retrieval, and composition.  
This separation limits narrative coherence, cross-modal stability, and fine-grained control in advertisement video creation.  
To address these limitations, we propose AutoCut, which discretizes video, audio, and text into a unified token vocabulary and performs end-to-end multimodal reasoning for coherent and controllable editing.
\vspace{-0.1in}

%% file: sec/3_method.tex
\section{Method}
\label{sec:method}
\vspace{-0.1in}

We propose AutoCut, a unified framework for multimodal discretization and controllable video editing.  
As illustrated in Figure~\ref{fig:Method}, AutoCut converts continuous video and audio embeddings into discrete tokens, trains a large language model to capture cross-modal dependencies, and reconstructs coherent short-video advertisements through token retrieval and rendering.
\vspace{-0.05in}
\subsection{Task Formulation}
\vspace{-0.05in}

Our framework is designed to handle multiple tasks in advertisement video editing, where multimodal inputs—text, video, and audio—interact within a shared representation space.  
Let $\mathcal{X} = \{x_p, x_t, x_v, x_a\}$ denote the set of possible input modalities, 
including product information ($x_p$), textual script ($x_t$), video clips ($x_v$), 
and background music ($x_a$).  
Depending on the specific task, the model receives a subset of these modalities as input 
and produces task-dependent outputs $y$, such as clip sequences, textual scripts, 
or background music.  

To comprehensively evaluate the framework, we define four representative tasks 
that correspond to the major stages of the ad-creation workflow: 
(1) video selection, (2) temporal ordering, (3) script generation, 
and (4) background music selection.  
These tasks together assess the model’s multimodal reasoning ability 
and its capacity to achieve coherent and controllable advertisement editing.

\vspace{-0.05in}
\subsection{Multimodal Encoder}
\vspace{-0.05in}
\noindent\textbf{Visual Encoder.}
We adopt an off-the-shelf CNN encoder based on the ResNet-50\cite{heDeepResidualLearning2016a} architecture to extract dense frame-level embeddings from video frames. The visual encoder was pre-trained with a contrastive learning objective on a large corpus of real-world advertising video frames. As a result, the extracted embeddings exhibit strong semantic expressiveness and high discriminability, which greatly facilitate subsequent quantization and enable effective retrieval of video segments in downstream applications. We choose this encoder as a practical trade-off between representation quality, cost, and efficiency.

\vspace{3pt}\noindent\textbf{Audio Encoder.}
We utilize Pretrained Audio Neural Networks (PANNs)~\cite{kongPANNsLargeScalePretrained2020a} 
to obtain high-level audio embeddings from each clip. 
Trained on the large-scale AudioSet dataset, PANNs demonstrate strong generalization 
across diverse acoustic environments. 
Each audio segment is first converted into a log-mel spectrogram and processed by 
the Wavegram-Logmel-CNN backbone to capture both temporal and spectral features. 
The resulting embeddings, representing the semantic acoustic content, are then 
discretized through the RQVAE quantizer to form audio tokens aligned with the 
unified multimodal vocabulary.

\subsection{Tokenization and Codebook Construction}

The video and audio embeddings are continuous, while large language models operate on discrete token sequences.  
To bridge this gap, we adopt the \textbf{Residual Quantized Variational AutoEncoder (RQVAE)}~\cite{lee2022rqvae} to discretize continuous multimodal representations.

RQVAE progressively approximates high-dimensional latent features through residual quantization, enabling efficient compression into a limited codebook.  
This process establishes a bidirectional mapping between continuous embeddings $f$ and discrete token indices $z$, trained via cosine reconstruction loss:
\begin{equation}
\mathcal{L}_{\mathrm{rec}} = 1 - \cos(\hat{f}, f),
\end{equation}
where $\hat{f}$ denotes the reconstructed feature.  
We set the codebook size to $256{\times}8$, encoding each video frame or audio segment into eight tokens.  
This configuration achieves high reconstruction quality (cosine similarity of 0.89 for video and 0.96 for audio) while maintaining manageable token length.  
The discrete tokens are then integrated into a shared multimodal vocabulary with text tokens.

\subsection{Multimodal Alignment}

With the unified token vocabulary established, we perform a multimodal alignment stage to teach the model how video, audio, and text tokens correspond to each other.  
Each training sample consists of temporally aligned sequences of multimodal tokens serialized into a unified input string.  
The model is optimized under a next-token prediction paradigm: 
\begin{equation}
\mathcal{L}_{\mathrm{NTP}} = - \sum_{t} \log P(x_t \,|\, x_{<t}),
\end{equation}
where $x_t$ denotes tokens from any modality.

We use Qwen3-8B~\cite{qwen3} as the base model and train on approximately 700K filtered advertisement samples.  
During alignment, the backbone is kept frozen, and only the newly introduced multimodal embedding layers are updated.  
This design stabilizes early training and ensures that video, audio, and text representations are projected into a shared semantic space before full task-specific adaptation.

Through this stage, the model learns essential cross-modal correspondences, such as script–scene grounding, audio–visual rhythm synchronization, and basic temporal structures common in short-video advertisements.

\subsection{Supervised Fine-Tuning}

Following multimodal alignment, we perform supervised fine-tuning (SFT) to teach the model task-specific behavior.  
In contrast to the alignment stage, which applies next-token prediction to the entire sequence, SFT computes the loss only over the response portion of the sequence.  
Each training sample consists of an input context and a designated target output such as an ordered clip sequence or a generated script. The model is optimized to accurately produce this target segment.

We employ full-parameter fine-tuning using the same optimization setup as in the alignment stage.  
Through this SFT procedure, AutoCut learns to perform controllable editing actions conditioned on user inputs, enabling coherent and stylistically consistent advertisement creation across a wide range of editing scenarios.

\subsection{Retrieval and Rendering}

After the model generates discrete token sequences, we reconstruct playable videos through retrieval and rendering as shown in Figure \ref{fig:Teaser}.  
Video tokens are mapped to the nearest clip embeddings in the material database using \textbf{FAISS}-based similarity search, producing visually matched scenes.  
Audio tokens are decoded or retrieved to align background music and voice-over components.  
Finally, \texttt{ffmpeg} is used to concatenate video clips, add transitions, overlay subtitles, and render the final MP4 output.  
This process bridges token-level editing decisions and executable video production, ensuring that AutoCut’s outputs are realistic, coherent, and production-ready.

\begin{figure*}[t]
  \centering
  \includegraphics[width=\linewidth]{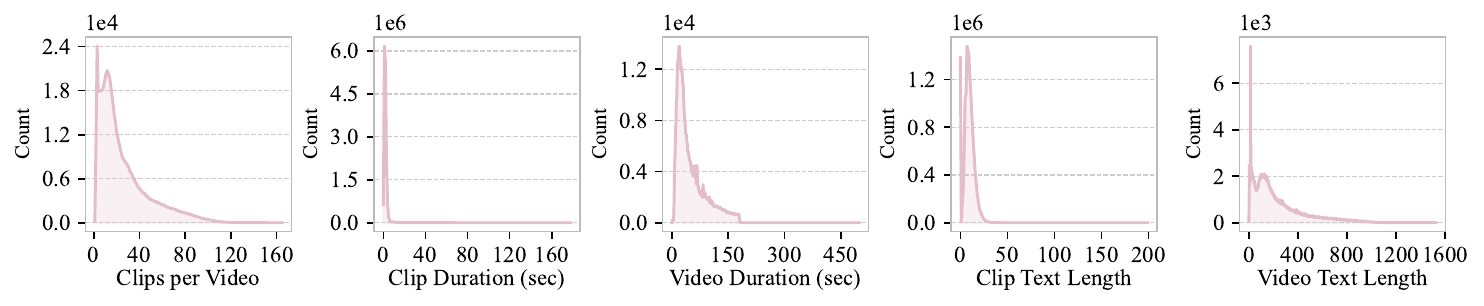}
  \vspace{6pt} 
  \includegraphics[width=\linewidth]{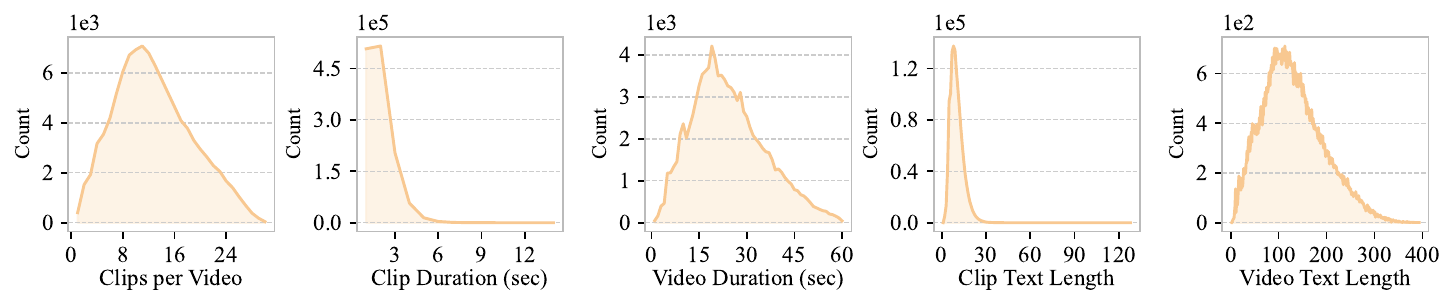}
  \caption{
Dataset statistics for the multimodal alignment (top) and SFT (bottom) stages. 
The alignment dataset is substantially larger but exhibits more diverse and irregular distributions, 
while the SFT dataset is smaller yet more balanced and of higher annotation quality.
}
  \label{fig:dataset_stats}
\vspace{-0.2in}
\end{figure*}

%% file: sec/4_exp.tex
\section{Experiments}
\label{sec:exp}

\subsection{Dataset Construction}

All data are collected from short-form advertisement videos sourced from online platforms.  
Each record includes a complete advertisement video together with its associated product metadata, such as category, brand, and selling points.

\vspace{3pt}\noindent\textbf{Data Parsing.}





Each advertisement video is parsed into three modalities: text, video, and audio. The spoken script is extracted by ASR and aligned with timestamps at the clip level. Video frames are sampled at 1 fps with \texttt{ffmpeg} and encoded into visual embeddings, while raw audio is separated with \texttt{pydub}, followed by speech--music separation and audio embedding extraction using \textit{pann\_inference}~\cite{kongPANNsLargeScalePretrained2020a}. Finally, each video is segmented into short clips according to punctuation-aligned ASR timestamps, yielding a sequence of synchronized multimodal segments.

\vspace{3pt}\noindent\textbf{Data Processing.}
The parsed multimodal data are used to construct two datasets: a large-scale multimodal alignment dataset and a smaller SFT dataset.  

For the multimodal alignment stage, advertisements with strong user engagement are retained, while videos that contain lyrical music or lack human speech are removed.  
Video and audio embeddings are discretized into token representations through RQVAE models, and the three modalities are integrated into unified advertisement samples.  
After filtering, the alignment dataset contains approximately 700K samples.

For the SFT stage, additional selection is applied to build a curated subset for downstream training.  
We retain videos shorter than 120 seconds, with clip lengths between 2 and 60 seconds, and with high visual-text relevance measured by Qwen-VL.  
Around 100K samples remain after filtering.  
Each record includes product metadata and is reorganized to support four task types:

\begin{itemize}
    \item \textbf{Video Selection:} Select relevant clips from a candidate pool.
    \item \textbf{Video Sorting:} Arrange selected clips into a coherent temporal sequence.
    \item \textbf{Script Generation:} Produce textual narration aligned with the visual content.
    \item \textbf{Background Music Selection:} Retrieve or generate music that matches the multimodal context.
\end{itemize}

This task-oriented design enables the model to learn multimodal understanding and prediction in a unified manner and forms the basis for controllable video editing.  
Detailed construction procedures are provided in the supplementary material.

\vspace{3pt}\noindent\textbf{Data Statistics.} Figure~\ref{fig:dataset_stats} reports the core statistics shared by both training stages, including the number of clips per video, clip duration, overall video duration, clip text length, and video text length.  
These measurements describe the temporal structure and linguistic density of advertisement videos.

For the multimodal alignment stage, we use a large-scale corpus that presents broad variation across all five metrics.  
Although the data contain imperfect boundaries and partially noisy transcripts, the scale and diversity of the corpus are valuable for learning stable cross-modal alignment among video, audio, and text tokens.

For the SFT stage, we curate a smaller yet significantly higher-quality subset.  
The same statistics apply, but the data provide cleaner segmentation and more accurate textual annotations, making them better suited for training controllable editing behaviors and coherent advertisement generation.

By combining a diverse alignment corpus with a curated SFT dataset, we adopt a two-stage pipeline consisting of multimodal alignment followed by supervised fine-tuning.  
This strategy removes the need for full-scale pretraining while maintaining strong performance.

\begin{table*}[t]
\centering
\setlength{\tabcolsep}{9pt}
\caption{Quantitative results across all tasks. The best results are boldfaced and the second best results are underlined.}
\label{tab:all-results}
\begin{tabular}{lccc ccc c}
\toprule
\multirow{2}{*}{Model} &
\multicolumn{3}{c}{Video} &
\multicolumn{2}{c}{Script} &
\multicolumn{1}{c}{Audio} \\
\cmidrule(lr){2-4} \cmidrule(lr){5-6} \cmidrule(lr){7-7}
& CSA & CRA & VSC & SQ & WCD $\downarrow$ & MSS \\
\midrule
Qwen3--8B\cite{qwen3technicalreport} (Caption)               & 0.1374 & 0.01648 & 0.9308 & 79.9947 & \underline{5.2648}  & -- \\
Qwen3--8B (Caption + SFT)               & 0.5687 & 0.03022 & \underline{1.1227} & 59.1592 & 6.8230  & -- \\
Qwen3--32B \cite{qwen3technicalreport}(Caption)              & 0.1731 & 0.01373 & 0.9342 & 80.7747 & 7.1023  & -- \\
Qwen2.5-VL--7B Instruct~\cite{Qwen2.5-VL} & 0.2418 & 0.01648 & 0.9649 & 79.4867 & 6.4709  & -- \\
Qwen2.5-VL--32B Instruct\cite{Qwen2.5-VL}          & \textbf{0.6648} & 0.02472 & 0.9980 & 78.3180 & 12.5071 & -- \\
InternVL-3.5-8B~\cite{wang2025internvl3_5} & 0.0247     & 0.01653      & 0.9466     & 81.0918      & 7.6415      & -- \\
LLaVA-v1.6-Mistral-7B-HF~\cite{llava} & 0.0027 & 0.00923 & 0.9914 & 56.1242 & 12.6580 & -- \\
GPT-4o~\cite{openaiGPT4oSystemCard2024} + MGSV~\cite{xin2025mgsv}  & 0.2689 & \underline{0.07756} & \textbf{1.1364} & \underline{83.0290} & 7.7457 & \underline{0.2656} \\
\midrule
\textbf{Autocut (ours) }                   & \underline{0.6593} & \textbf{0.10714} & 1.0360 & \textbf{84.6255} & \textbf{3.0182} & \textbf{0.3475} \\
\bottomrule
\end{tabular}
\vspace{-0.1in}
\end{table*}

\subsection{Evaluation Metrics}

We employ a set of quantitative and qualitative metrics to evaluate different aspects of the proposed framework, including multimodal alignment, video clip selection, ranking consistency, script quality, and audio retrieval performance.  
All metrics are defined below.

\vspace{3pt}\noindent\textbf{Visual–Script Correlation (VSC).}
VSC measures the semantic consistency between the selected video clips and the corresponding advertisement script.  
Each video–script pair is evaluated by a large language model (GPT-4o), which assigns a discrete score from $\{0, 1, 2\}$ based on the degree of alignment, where higher values indicate stronger semantic correspondence.

\vspace{3pt}\noindent\textbf{Clips Selection Accuracy (CSA).}
CSA quantifies the accuracy of identifying positive video clips.  
It is defined as:
\[
\text{CSA} = \frac{N_{\text{select}}}{N_{\text{total}}} \times 100\%,
\]
where $N_{\text{select}}$ denotes the number of samples that contain only positive clips, and $N_{\text{total}}$ represents the total number of evaluated samples.

\vspace{3pt}\noindent\textbf{Clips Rank Accuracy (CRA).}
CRA evaluates the correctness of the predicted order of video clips.  
It is calculated as:
\[
\text{CRA} = \frac{N_{\text{correct}}}{N_{\text{total}}} \times 100\%,
\]
where $N_{\text{correct}}$ is the number of samples whose predicted clip sequences match the reference ordering exactly, and $N_{\text{total}}$ is the total number of samples.

\vspace{3pt}\noindent\textbf{Script Quality (SQ).}
SQ evaluates the overall textual quality of the generated advertisement scripts.  
We employ a GPT-4o–based evaluator following a three-category, 100-point rubric that covers:  
(1) \textit{Basic Quality} (30 pts), assessing factual correctness, clarity, and content safety;  
(2) \textit{Expression and Communication} (40 pts), measuring linguistic naturalness, audience engagement, and the clarity of selling-point delivery;  
and (3) \textit{Length and Rhythm} (30 pts), including line-level length consistency with reference scripts and the fluency of phrasing and pausing.  
Scores from these categories are summed to obtain the final SQ score.


\vspace{3pt}\noindent\textbf{Clip-Level Word Count Discrepancy (WCD).}
To evaluate the temporal consistency between the generated script and each video clip, we introduce the clip-level Word Count Discrepancy:
\[
WCD =
\left|
\text{WordCount}_{\text{script}} -
\text{WordCount}_{\text{target}}
\right|.
\]
A lower WCD indicates closer alignment between the textual content and the temporal structure of the corresponding video clip, ensuring that the generated script matches the expected speaking duration and maintains temporal consistency at the clip level.


\vspace{3pt}\noindent\textbf{Music Similarity Score (MSS).}
To evaluate the quality of background music prediction, we measure the semantic similarity between the predicted BGM and the ground-truth BGM.
For each audio pair $(a_{\text{pred}}, a_{\text{gt}})$, we use the Music Flamingo~\cite{ghosh2025musicflamingoscalingmusic} model to first generate structured descriptions of both audio clips (covering genre, tempo, instrumentation, structure, and emotional tone).
The model then compares the two descriptions and outputs a continuous similarity score in the range $[0,1]$:
\[
\text{MSS} =
\text{FlamingoSim}\!\left(
\text{Desc}(a_{\text{pred}}),\,
\text{Desc}(a_{\text{gt}})
\right).
\]
A higher score indicates that the predicted background music is semantically closer to the ground-truth audio in terms of musical attributes and emotional expression.

\subsection{Experimental Details}
The training parameter settings are provided in the supplementary materials. For evaluation, we adopt a unified evaluation protocol across all four tasks. All models are prompted using the same
instruction templates to ensure consistent task definitions and comparable output formats. All
experiments are conducted on the same split of 364 videos to guarantee fairness across model types.

For text-only models such as Qwen3-8B and Qwen3-32B, which cannot directly process visual inputs, each video clip is converted into a short caption using Qwen2.5-VL-32B. The caption describes the clip's first frame and serves as a lightweight textual proxy for the visual content, allowing text-only models to participate in all video-related tasks without introducing additional modality gaps. 
We additionally fine-tune a Qwen3-8B (Caption) baseline on the same SFT dataset used by AutoCut. This baseline uses the same task definitions and evaluation split, but operates on captionized clip inputs rather than discrete video tokens.
Multimodal models retain their native frame-based visual inputs. AutoCut further operates on discretized low-fps video tokens and optionally incorporates textual or audio tokens depending on the task, following its unified multimodal representation scheme. We also introduce powerful closed-source multimodal models such as GPT-4o as our baseline to demonstrate the effectiveness of our approach.

For the background music retrieval task, we benchmark only against MGSV~\cite{xin2025mgsv} with GPT-4o, which is the sole baseline equipped with a native audio encoder and music–video matching capability. 

Full prompt templates and example instructions are provided in the Supplementary Material.




\subsection{Experimental Results}

Table~\ref{tab:all-results} summarizes the quantitative results on all four tasks. The comparison includes both off-the-shelf text-only and multimodal LLM baselines, as well as a Qwen3-8B (Caption) baseline trained on the same SFT tasks.

Across all tasks, AutoCut demonstrates strong and well-rounded performance. It obtains the best CRA score of 0.10714, indicating a clear advantage in recovering correct clip order and modeling temporal structure. AutoCut also achieves the best SQ score of 84.6255, the lowest WCD of 3.0182, and the highest MSS of 0.3475, showing superior script quality, temporal coherence, and background-music matching. On clip-level understanding, it achieves a CSA score of 0.6593, comparable to the best-performing vision–language baselines, despite relying on lightweight discretized video tokens.

Although GPT-4o and the fine-tuned Qwen3-8B (Caption) baseline both slightly outperform AutoCut on VSC, AutoCut still achieves a strong score of 1.0360, indicating good script--visual correspondence. 
More importantly, VSC mainly reflects local semantic consistency rather than overall editing quality. AutoCut remains stronger on the more practical task-level metrics, including CSA, CRA, SQ, and WCD, while also being substantially more efficient than caption-based pipelines.

In addition to strong editing performance, AutoCut is also substantially more cost-efficient in deployment. Based on our estimate, processing 100 videos costs about \$2.5 with a GPT-4o API pipeline, compared with about \$0.015 for AutoCut on a single RTX 4090, showing an order-of-magnitude reduction in inference cost.

Taken together, these results show that AutoCut provides the best overall balance across the key dimensions of advertisement video editing. Its token-based multimodal representation is particularly effective for structured editing tasks, leading to stronger temporal reasoning, better script planning, and higher practical usability. 

\begin{table*}[t]
\centering
\small
\setlength{\tabcolsep}{10pt}
\renewcommand{\arraystretch}{1.12}
\caption{Ablation on training settings.}
\label{tab:ablation-narrow}
\begin{minipage}{0.90\textwidth}   
\centering
\begin{tabular}{lccccc}
\toprule
Method & CSA & CRA & VSC & SQ & WCD $\downarrow$ \\
\midrule
sft only        & 0.4780 & 0.08242 & 1.0043 & 83.1898  & 4.4346   \\
emb+full+sft     & \textbf{0.7170} & 0.05770 & 0.9669 & 78.9644 & 4.4984 \\
emb+sft (ours)         & 0.6593 & \textbf{0.10714} & \textbf{1.0360} & \textbf{84.6255}  & \textbf{3.0182}      \\
\bottomrule
\end{tabular}
\end{minipage}
\vspace{-6pt}
\end{table*}

\begin{figure}[t]
  \centering
   \includegraphics[width=\linewidth]{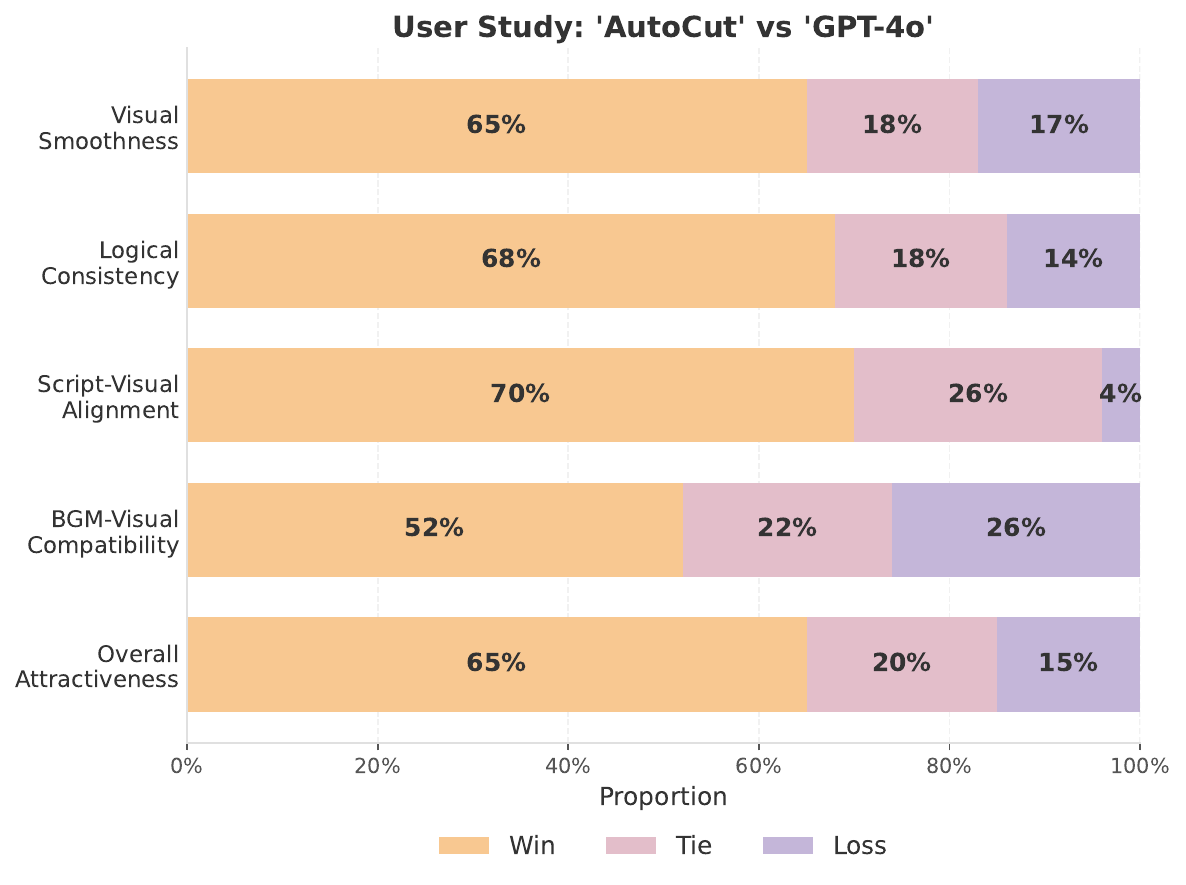}
   \vspace{-0.2in}
  \caption{
User study results: win–loss ratios against GPT-4o across five evaluation dimensions.
}
  \label{fig:user_study}
\end{figure}

\subsection{Human Evaluation Results}
Given the difficulty of fully assessing advertisement video quality through automated metrics alone, we further conducted a human evaluation to measure perceptual preference. We performed a pairwise comparison study between videos created by AutoCut and those produced by the strongest multimodal LLM baseline, GPT-4o. We recruited 10 independent annotators and curated a set of 100 representative test samples. For each sample, the annotators were shown two completed advertisement videos (A and B), both promoting the same product and constructed from the same pool of raw candidate clips.

After watching both videos, annotators were asked to indicate their preference across five perceptual dimensions: visual smoothness, logical consistency, script-visual alignment, BGM-visual compatibility, and overall advertisement appeal. For each dimension, participants could choose \textit{Win}, \textit{Loss}, or \textit{Tie} when the two versions were perceived as equally good.

We summarize the preference trends in each dimension by aggregating the proportion of times each method was favored. As shown in Figure~\ref{fig:user_study}, our method consistently outperforms the strong GPT-4o baseline across all evaluation dimensions, particularly in Script-Visual Alignment, highlighting the benefits of our multimodal joint modeling.

Due to space constraints, additional qualitative visual results can be found in the supplementary material.

\subsection{Ablations}
To validate the effectiveness of our two-stage training pipeline, we conduct ablations across three settings (Table~\ref{tab:ablation-narrow}): (1) \textit{sft only}, (2) \textit{emb+full+sft}, and (3) \textit{emb+sft} (ours). Our framework is designed as a lightweight two-stage process: the first stage performs multimodal alignment, and the second stage applies SFT to learn task-specific editing behavior.

\vspace{3pt}
\noindent\textbf{Effect of Multimodal Alignment.}
Compared with \textit{sft only}, SFT introducing the alignment stage (\textit{emb+sft}) yields consistent improvements across CSA, CRA, VSC, and SQ, while significantly reducing WCD. This confirms that explicitly aligning text, audio, and visual tokens provides a stronger initialization for downstream editing tasks.

\vspace{3pt}
\noindent\textbf{Effect of Additional Pre-training.}
We further evaluate an additional pre-training stage after multimodal alignment (\textit{emb+full+sft}). Despite slightly improving CSA, this setting degrades CRA and SQ and increases WCD. We attribute this behavior to the limited quality and weak label consistency of the available pre-training corpus, which introduces noise into the learned token distributions and interferes with the alignment stage. Therefore, we adopt the two-stage \textit{emb+sft} pipeline as our default training strategy, as it provides the best overall balance of accuracy, consistency, and editing stability.

%% file: sec/5_discussion.tex
\section{Discussion}
\textbf{Limitations.}
Although AutoCut demonstrates strong performance across multimodal understanding and generation tasks, several limitations remain. 
(1) While unified tokenization substantially improves cross-modal alignment, subtle desynchronization between video motion and audio rhythm still occurs, indicating that multimodal coupling is not yet fully unified. 
(2) Controllability remains primarily at the clip level; the current framework does not yet support fine-grained, emotion-aware, or frame-level editing, which limits narrative flexibility. 
(3) AutoCut adopts a material-based editing paradigm, in which the rendering stage retrieves and composes existing video and audio assets from a material database. This design improves realism and production readiness, but it does not synthesize new video pixels or raw audio from scratch, and therefore remains bounded by the coverage and quality of the available materials.

\noindent \textbf{Future Directions.}
Future work will address these limitations from multiple perspectives. 
(1) Improve temporal alignment and multimodal coherence through cross-attentive modeling and diffusion-based refinement, achieving tighter video--audio--text synchronization. Future work can also explore stronger visual front-ends, such as ViT, CLIP, or video-based encoders, to further improve representation quality.
(2) Enhance controllability via expanded instruction sets and latent-space editing, enabling the model to manipulate rhythm, tone, and emotion at a finer level. 
(3) Integrate learned clip synthesis and adaptive retrieval into a single decoding process, unifying generation and rendering and moving AutoCut toward fully generative, human-aligned, and scalable video creation.

%% file: sec/6_conclusion.tex
\vspace{-0.1in}
\section{Conclusion}
\label{sec:con}

This paper presents AutoCut, an end-to-end framework for advertisement video editing based on multimodal discretization and controllable editing. By transforming video, audio, and textual inputs into a shared discrete token space and aligning them through large-scale multimodal pretraining and fine-tuning, AutoCut bridges perception and creation within a unified reasoning framework. Extensive experiments on real-world advertisement datasets demonstrate that AutoCut achieves superior semantic coherence, controllability, and production efficiency compared with retrieval- or template-based baselines. The study establishes multimodal tokenization as a foundation for connecting discrete reasoning and generative synthesis in video editing, paving the way toward fully generative, human-aligned, and scalable AI-driven media production.

%% file: sec/7_suppl.tex
\clearpage
\setcounter{page}{1}
\maketitlesupplementary

\section{Appendix}
\label{sec:appendix}

This appendix is organized as follows:
\begin{itemize}
  \item \textbf{7.1 Model Architecture and Hyperparameters} \\
  model components and key training settings.
  \item \textbf{7.2 Post-Processing and Rendering Pipeline} \\
  steps for converting predictions into final ad videos.
  \item \textbf{7.3 GPT-4o Evaluation Prompts} \\
  prompts used for GPT-4o-based automatic evaluation.
  \item \textbf{7.4 Human Evaluation Criteria} \\
  criteria and guidelines for the pairwise user study.
  \item \textbf{7.5 Dataset Construction Details} \\
  details of alignment data, SFT filtering, and task formatting.
\end{itemize}

\subsection{Model Architecture and Hyperparameters}

\begin{table}[h]
\centering
\setlength{\tabcolsep}{3pt}
\caption{Hyperparameters for the video and audio RQ-VAE quantizers.}
\label{tab:rqvae-hparams}
\begin{tabular}{lcc}
\toprule
\textbf{RQ-VAE Configs} & \textbf{Video} & \textbf{Audio} \\
\midrule
Input feature size       & 128                   & 2048                  \\
Quantization heads       & 8                     & 8                     \\
Codebook size            & 256                   & 256                   \\
Codebook dim             & 128                   & 256                   \\
Shared codebook          & False                 & False                 \\
Encoder MLP size         & [512, 512]            & [1024, 512]           \\
Decoder MLP size         & [512, 512]            & [512, 1024]           \\
Loss type                & cosine                & cosine                \\
Reconstruction weight    & 1.0                   & 1.0                   \\
Commitment weight        & 0.0                   & 0.0                   \\
Optimizer weight decay   & 1e$^{-4}$             & 1e$^{-4}$             \\
Initial learning rate    & 1e$^{-3}$             & 1e$^{-3}$             \\
Min learning rate        & 2e$^{-5}$             & 2e$^{-5}$             \\
Batch size               & 8192                  & 8192                  \\
Max epochs               & 20                    & 50                    \\
\bottomrule
\end{tabular}
\end{table}

To convert continuous visual and audio embeddings into discrete tokens, we train two modality-specific Residual Quantized Variational Autoencoders (RQ-VAEs). Both models use multi-head residual quantization, cosine reconstruction loss, and large-batch optimization, while differing in encoder--decoder capacity to match modality characteristics.

For \textbf{video}, 128-dimensional frame embeddings are quantized using an 8-head residual quantizer with a 256-entry codebook and lightweight two-layer MLP encoder--decoder. This configuration balances reconstruction fidelity and token compactness for downstream selection and ordering tasks.

For \textbf{audio}, the 2048-dimensional PANNs embeddings require a wider asymmetric MLP architecture, while retaining the same quantizer heads and codebook size. Training follows the same objective and optimizer settings but uses a longer schedule to accommodate higher input dimensionality.

Both quantizers are trained on large-scale advertisement datasets and monitored with Weights \& Biases for stability. A consolidated comparison of all model hyperparameters is provided in Table~\ref{tab:rqvae-hparams}.

\begin{table}[t]
\centering
\setlength{\tabcolsep}{3pt}
\caption{Hyperparameters used during the multimodal alignment and SFT stages.}
\label{tab:align-sft-hparams}
\begin{tabular}{lcc}
\toprule
\textbf{Config} & \textbf{Alignment Stage} & \textbf{SFT Stage} \\
\midrule
Base model & Qwen3-8B & Aligned model\\
Template & qwen3 & qwen3 \\
Cutoff length & 8096 & 4000 \\
Packing & Enabled & Disabled \\
Per-device batch size & 2 & 1 \\
Grad. accumulation & 1 & 1 \\
Learning rate & 5e--5 & 1e--5 \\
Epochs & 5 & 3 \\
LR scheduler & Cosine & Cosine \\
Warmup ratio & 0.1 & 0.1 \\
Precision & bf16 & bf16 \\
Optimizer & AdamW & AdamW \\
Flash attention & fa3 & fa3 \\
Resize vocab & Enabled & Disabled \\
\bottomrule
\end{tabular}
\end{table}

We adopt a two-stage training pipeline to integrate multimodal discrete tokens into the Qwen3-8B backbone and enable controllable editing. The first stage performs \textbf{multimodal alignment}, where the backbone is frozen except for the newly added multimodal embedding layers. This stage learns semantic correspondence across video, audio, and text tokens using a cosine learning-rate schedule, a long context window (cutoff 8096), and bf16 mixed precision. Training uses a per-device batch size of 2, no gradient accumulation, and runs for five epochs at a learning rate of 5e--5, with packing and DeepSpeed ZeRO-2 to improve memory and throughput.

\begin{figure*}[t]
  \centering
  \includegraphics[width=0.81\linewidth]{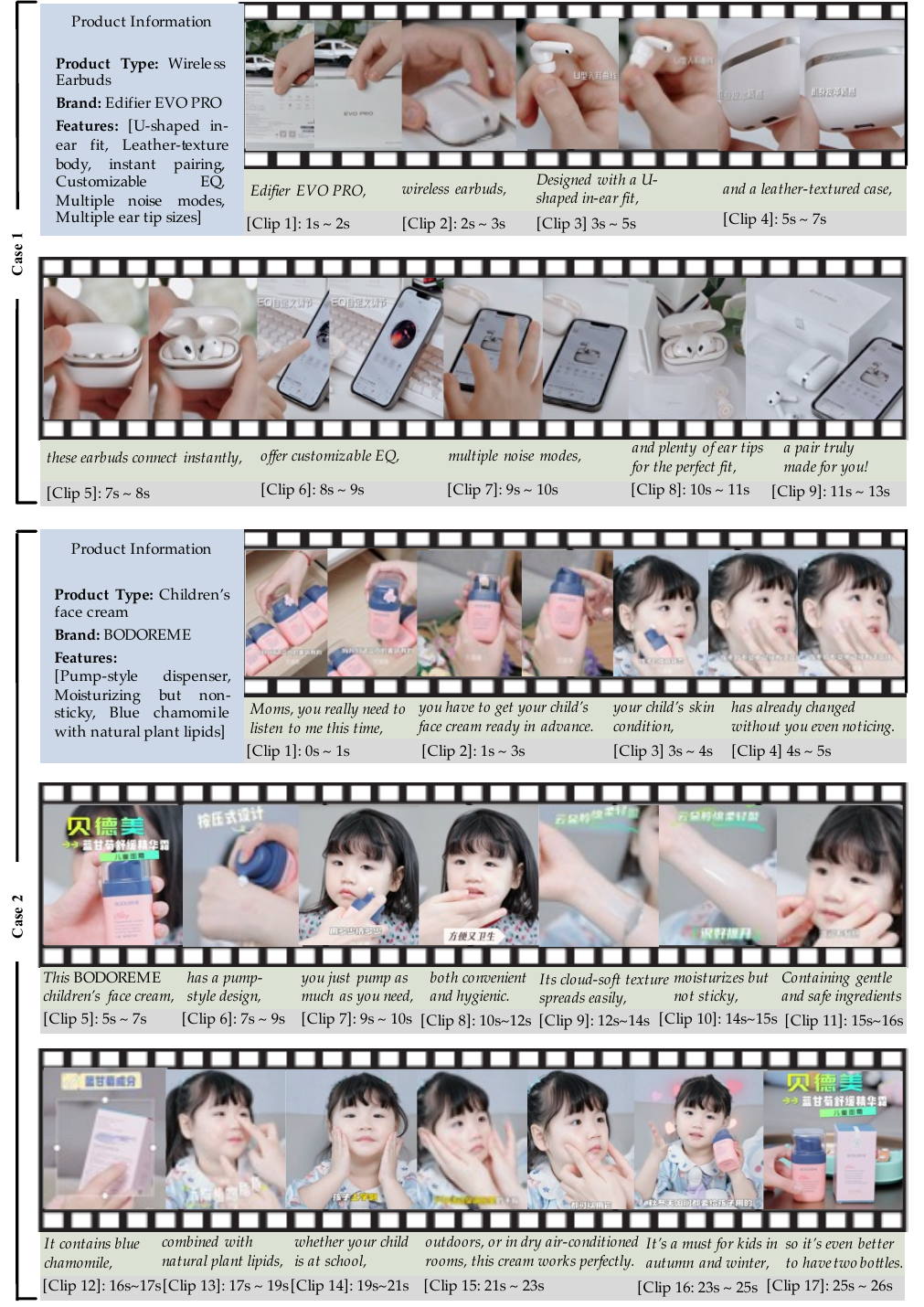}
\caption{
Qualitative case studies of our automatic ad video editing pipeline.
For each case (top: wireless earbuds, bottom: children’s face cream), we show the provided product information, selected video clips, and the aligned script sentences.
Each frame strip visualizes the model’s clip selection and ordering, together with the corresponding sentence and clip-level timestamps, illustrating how the system produces a coherent, time-aligned ad from raw footage and product metadata.
}

  \label{fig:placeholder}
\end{figure*}

The second stage applies \textbf{supervised fine-tuning (SFT)} on the aligned model. Unlike alignment, SFT updates all parameters and disables packing to preserve task structure. It is trained on curated datasets covering video selection, clip ordering, script generation, and BGM selection, using a shorter sequence length (cutoff 4000), a smaller learning rate (1e--5), and a per-device batch size of 1 for three epochs. The same optimizer, scheduler, and mixed-precision settings are used as in the alignment stage.

All training experiments for both stages are conducted on a cluster of 8 NVIDIA GPUs. All templates used during SFT follow the \texttt{qwen3} instruction format, and model checkpoints are logged using Weights \& Biases. Table~\ref{tab:align-sft-hparams} summarizes the hyperparameters used in the alignment and SFT stages.

\subsection{Post-Processing and Rendering Pipeline}
\label{sec:postprocess}

In the main paper we have shown that our model achieves strong performance on all sub-tasks and outperforms comparable baselines, both in quantitative metrics and in our user study. Here, we detail the complete post-processing and rendering pipeline that converts model outputs into a final edited video.
For clarity, AutoCut does not perform clip or BGM selection by directly predicting fixed database indices. Instead, the model outputs discrete token sequences that are decoded into target embeddings, and the final video or audio assets are retrieved from the material database through similarity search. This retrieval-based design allows the system to generalize to unseen clips and music tracks, rather than memorizing a closed set of asset IDs.

\paragraph{Editing scenarios and model outputs.}
We consider two typical usage scenarios.

\textbf{(1) Script–driven editing.}
The user provides (i) product information, (ii) a pool of candidate product clips, and (iii) a target ad script.
\emph{Selection.} Given the script and clip pool, the model first performs a \emph{select} task: it chooses clips that are relevant to the script. We constrain the number of selected clips to match the number of sentences in the script.
\emph{Sorting.} The model then performs a \emph{sort} task on the selected clips, ordering them such that visual content aligns with the sentence-level script order, based on learned vision–language correlations.
\emph{BGM selection.} Finally, we feed the ordered clip sequence together with the script and product information into the model for \emph{background-music (BGM) selection}. The model outputs a sequence of discrete audio tokens corresponding to a suitable BGM track for this rough cut.

\textbf{(2) Footage–driven editing.}
In the second scenario, the user has already curated all relevant footage but has not yet written an ad script; effectively, the \emph{select} step has been performed manually.
Given the product information and the set of clips, the model first \emph{generates an ad script} whose number of sentences matches the number of clips.
We then apply the same \emph{sort} and \emph{BGM selection} steps as in the script–driven pipeline.

In both scenarios, the model outputs discrete video tokens (for clips), audio tokens (for BGM), and the text script.

\paragraph{Token decoding and retrieval.}
To map token sequences back to concrete media assets, we use the RQVAE models trained in the pre-processing stage. We first decode the discrete tokens into continuous embeddings, and then perform nearest-neighbor search in FAISS indices.

For \textbf{video tokens}, FAISS returns a pair \((\texttt{frame\_id}, \texttt{clip\_id})\) for each token. These identifiers are structured:
\begin{itemize}
  \item \texttt{frame\_id} is constructed from a base \texttt{photo\_id} (the underlying source video) plus four digits encoding the frame index within that video.
  \item \texttt{clip\_id} is constructed from the same \texttt{photo\_id} followed by seven digits, where the first four digits denote the starting frame of the clip and the last three digits encode its duration (at 1\,fps).
\end{itemize}
For \textbf{audio tokens}, FAISS returns an \texttt{audio\_id} pointing to a BGM track in our library.

\paragraph{Rendering strategies: by frame vs.\ by clip.}
Using these identifiers, we support two strategies for assembling the final video.

\textbf{By frame.}
We derive the starting frame from the last four digits of \texttt{frame\_id} and use the stored clip length (in frames at 1\,fps) to cut a contiguous segment from the original video. This provides frame-accurate control over clip boundaries.

\textbf{By clip.}
We additionally maintain an alternative clip segmentation defined independently of script boundaries. Each source video is segmented into visually coherent clips based on changes in visual continuity: we detect boundaries where the similarity between consecutive frame embeddings drops sharply. For each such segment (computed at 20\,fps), we average the frame embeddings to obtain a single clip embedding. At inference time, retrieval is performed over these ``visual clips'', and we stitch together the corresponding segments.
This \emph{by-clip} strategy is specifically designed to reduce visual jitter or abrupt cuts that may occur if we follow sentence boundaries alone without considering the intrinsic visual structure of the footage.

\paragraph{Subtitles, TTS, and BGM mixing.}
For each clip in the final sequence, we associate one sentence of the provided or generated script. We render the sentence as on-screen subtitles and synthesize a voice-over track using text-to-speech (TTS). In the current implementation we use a fixed ``avatar'' voice; extending the model to select an appropriate voice profile conditioned on the video content is left for future work.
Finally, we retrieve the BGM track using the predicted \texttt{audio\_id} and mix it with the TTS audio when compositing the final video.

For all user-study videos and the demo results reported in the paper, we adopt the \textbf{script–driven} editing pipeline (given script) together with the \textbf{by-clip} rendering strategy.

\subsection{GPT-4o Evaluation Prompts}

We provide additional details on the automated evaluation procedures used to measure semantic alignment and script quality. Full prompts are included as separate text files; here we summarize their intended behavior.

\paragraph{VSC Scoring Prompt}
The Visual--Script Correlation (VSC) evaluator compares a single video frame with its corresponding script line. It is instructed to judge semantic consistency and output a discrete score in $\{0,1,2\}$, indicating mismatch, partial relevance, or strong alignment. The evaluator receives only the frame and the script line and is constrained to return a numerical score. The full prompt is provided in \texttt{vsc\_prompt.txt}.

\paragraph{SQ Scoring Prompt}
The Script Quality (SQ) evaluator applies a 100-point rubric covering correctness, clarity, linguistic naturalness, selling-point communication, and timing/length alignment. It is given product metadata, a human reference script, the generated script, and a pre-computed line-level length-difference statistic. The evaluator outputs a structured JSON object with category-level scores and a brief justification. The full prompt is included in \texttt{sq\_prompt.txt}.

To ensure consistent behavior across evaluation runs, both prompts explicitly define scoring boundaries, output formats, and constraints that prevent the evaluator from introducing additional commentary or unsupported interpretations.

\subsection{Human Evaluation Criteria}

To assess perceptual advertisement quality beyond automated metrics, we conducted a controlled user study with ten professional advertising specialists. Each participant evaluated ten video pairs, resulting in a total of 100 comparisons between videos generated by \textit{AutoCut} and by the \textit{GPT-4o + MGSV} baseline. For fairness, the ordering of the two videos in each pair was fully randomized, ensuring that annotators could not infer the model identity. Instead of collecting numerical scores, we adopt a pairwise comparison protocol in which annotators simply choose the better video for each criterion. This approach reduces calibration bias, avoids inconsistent use of rating scales, and yields more stable and discriminative perceptual judgments.

Annotators compared each video pair across five dimensions: visual smoothness, logical consistency, script–visual alignment, BGM–visual compatibility, and overall advertisement attractiveness. Table~\ref{tab:human-criteria} summarizes the guidelines used by annotators to judge the winner in each dimension.


\begin{figure}[t]
  \centering
  \includegraphics[width=\linewidth]{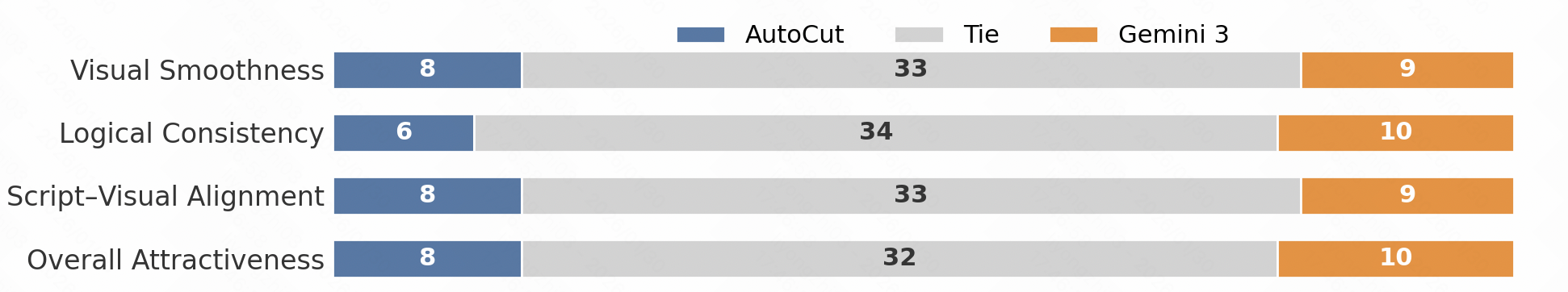}
  \caption{
  Additional human study comparing \textbf{AutoCut} with \textbf{Gemini 3} on 50 advertisement videos. A large proportion of cases are judged as \textit{Tie} across all five dimensions, indicating that AutoCut achieves performance highly comparable to this stronger and more recent baseline.
  }
  \label{fig:gemini_user_study}
\end{figure}

\subsection{Additional Human Study with Gemini 3}
\label{sec:gemini_human_study}

Since multimodal foundation models continue to improve rapidly, we further compare AutoCut with a stronger and more recent closed-source baseline, Gemini 3, which was released after our original submission. We conduct an additional pairwise human evaluation on \textbf{50} advertisement videos using the same protocol as the main user study.

For each sample, annotators compare the completed advertisement videos generated by AutoCut and Gemini 3 across four dimensions: \textit{visual smoothness}, \textit{logical consistency}, \textit{script--visual alignment}, and \textit{overall advertisement attractiveness}. As in the main study, annotators choose \textit{Win}, \textit{Loss}, or \textit{Tie} for each dimension.

Figure~\ref{fig:gemini_user_study} summarizes the results. We observe that \textbf{AutoCut} achieves highly comparable performance to Gemini 3, with a majority of cases resulting in \textit{Tie} across all evaluation dimensions. This result suggests that AutoCut remains competitive even against a much stronger and more recent multimodal baseline. Importantly, AutoCut achieves this performance with a substantially smaller parameter scale and lower computational cost, highlighting its practical efficiency for advertisement video editing.


\subsection{Dataset Construction Details}

\begin{figure*}[t]
  \centering
  \includegraphics[width=\linewidth]{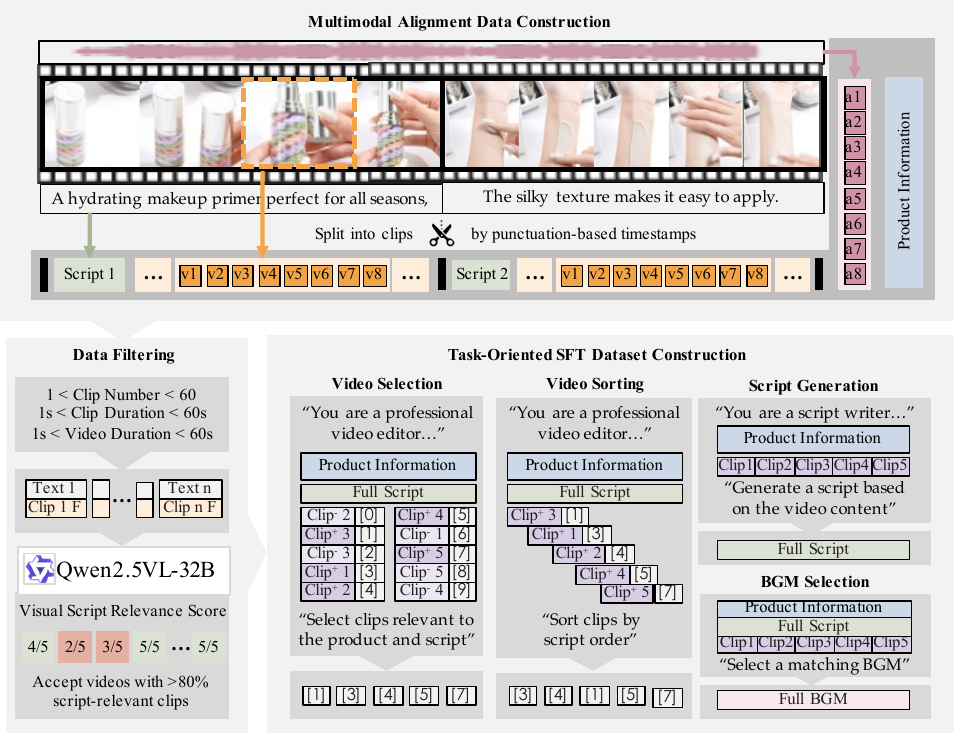}
\caption{
Overview of our dataset construction pipeline.
\emph{Top}: multimodal alignment data construction. Raw ad videos are paired with sentence-level scripts and product information, then segmented into short clips using punctuation-aligned timestamps, producing aligned video tokens and audio tokens.
\emph{Bottom-left}: data filtering. We keep videos whose clip number and duration fall within preset ranges and whose visual–script relevance score (estimated by Qwen2.5-VL-32B) exceeds an 80\% threshold.
\emph{Bottom-right}: task-oriented SFT dataset construction. From the filtered aligned data we instantiate four supervision signals: video selection, video sorting, script generation, and BGM selection, each conditioned on product information and/or the full script.
}
  \label{fig:data_construction}
\end{figure*}

\subsubsection{Multimodal Alignment Dataset Construction}
\label{sec:multi_align}

We start from short-form advertisements with high user engagement, defined as
videos whose click-through rate and like-rate both lie in the top 10\% on the
platform. For each video we collect product metadata (category, brand, selling
points) and process the three modalities in parallel
(Fig.~\ref{fig:data_construction}).

On the \textbf{video} side, we sample frames at 1\,fps with \texttt{ffmpeg} and
encode them using an internal CNNv2 to obtain dense visual embeddings. On the
\textbf{audio} side, we extract the audio track, separate background music from
speech, and embed the BGM with a PANNs-based service into 2048-D acoustic
features. On the \textbf{text} side, ASR is applied to the speech channel to
obtain time-stamped transcripts; a lightweight Qwen3-0.6B filter removes videos
whose transcripts mainly contain song lyrics or lack coherent product
descriptions.

Visual and acoustic features are then discretized by the RQ-VAE tokenizers
(Sec.~3.3) to produce sequences of video and audio code indices. We define clips
by ASR punctuation boundaries and \emph{soft-align} visual tokens whose
timestamps fall inside each sentence span. For every ad, clip-level text and
video tokens are concatenated into a single sequence, and the audio-token
summary of the full BGM track is appended. This yields our multimodal alignment
corpus for the first training stage, containing roughly \textbf{700K}
advertisements.

\subsubsection{SFT Data Filtering Pipeline}
\label{sec:sft_filter}

From the alignment set, we derive an SFT-ready subset of about
\textbf{100K} high-quality ads using the following filters:

\begin{itemize}
  \item \textbf{Duration:} keep videos shorter than 120\,s and clips within
  2--60\,s.
  \item \textbf{Caption quality:} discard ads with empty ASR, incoherent lines,
  or heavy repetition.
  \item \textbf{Visual–text relevance:} for each clip, pair the first frame with
  its ASR line and query Qwen2.5-VL-32B for a 0--5 relevance score; a video is
  retained only if at least \textbf{80\%} of its clips score $\geq 4$.
  \item \textbf{Deduplication:} remove near-duplicates by hashing
  \{\emph{brand, product, script}\}.
\end{itemize}

This pipeline enforces strong multimodal consistency and clean semantic grounding
before constructing the task-oriented SFT datasets.

\subsubsection{Task-Oriented SFT Dataset Construction}

The four supervised tasks introduced in Sec.~3 are constructed from the filtered pool.
For training convenience, some SFT tasks are expressed in an index-based output format. These indices serve only as task-level supervision over the local candidate pool. In the full AutoCut pipeline, however, final video and audio asset grounding is performed by embedding-based retrieval from the material database, as described in Sec.~\ref{sec:postprocess}.
Since the main text focuses on task definitions, here we summarize only the input--output formatting used to build the SFT dataset. All tasks are encoded in a \textbf{ShareGPT-style two-turn} format with a \verb|system| role and a single \verb|human|--\verb|assistant| exchange.

\begin{itemize}
  \item \textbf{Video Selection.}
  \textbf{Input:} product metadata, the full ad script, and a pool of candidate
  multimodal clips (each clip represented by its text snippet and video tokens).
  The clips are presented in random order, and we make sure that positive and negative clip ratios are 1:1.
  \textbf{Output:} the subset of candidate clips that should appear in the final ad, represented as indices within the local candidate pool.

  \item \textbf{Video Sorting.}
  \textbf{Input:} product metadata, the ordered reference script, and a set of
  selected clips whose order has been randomly shuffled.
  \textbf{Output:} an ordering of the selected candidate clips, represented as a permutation over local candidate indices.

  \item \textbf{Script Generation.}
  \textbf{Input:} product metadata and an ordered sequence of clips.
  \textbf{Output:} a multi-line ad script whose number of sentences matches the
  number of clips, providing a sentence-level description aligned with the
  visual sequence.

  \item \textbf{BGM Selection.}
  \textbf{Input:} product metadata, the ordered clips, and the full script.
  \textbf{Output:} a discrete audio-token sequence representing the target background-music style, which is later grounded to a concrete track through embedding-based retrieval.
\end{itemize}

For each task we construct approximately \textbf{25K--30K} instances, and adopt
a \textbf{95/5} train--validation split. The clip-level multimodal structure
and the end-to-end SFT data construction process are summarized in
Fig.~\ref{fig:data_construction}.


\begin{table*}[t]
\centering
\setlength{\tabcolsep}{4pt}
\renewcommand{\arraystretch}{1.2}
\caption{Human evaluation criteria and decision guidelines used in the pairwise comparison study.}
\label{tab:human-criteria}

\begin{tabular}{|p{0.17\textwidth}|p{0.35\textwidth}|p{0.35\textwidth}|}
\hline
\textbf{Criterion} & \textbf{Positive indicators} & \textbf{Negative indicators} \\
\hline

\parbox[t]{0.17\textwidth}{\centering \textbf{Visual}\\\textbf{Smoothness}} &
\begin{itemize}\setlength{\itemsep}{0pt}\setlength{\parsep}{0pt}\setlength{\topsep}{0pt}
  \item Transitions feel natural and continuous.
  \item No obvious stutter, jitter, or dropped frames.
  \item Motion and camera movement appear stable.
\end{itemize}
&
\begin{itemize}\setlength{\itemsep}{0pt}\setlength{\parsep}{0pt}\setlength{\topsep}{0pt}
  \item Noticeable lag, stutter, or frame skipping.
  \item Abrupt cuts that break visual flow.
  \item Shaky or inconsistent motion that distracts the viewer.
\end{itemize}
\\ \hline

\parbox[t]{0.17\textwidth}{\centering \textbf{Logical}\\\textbf{Consistency}} &
\begin{itemize}\setlength{\itemsep}{0pt}\setlength{\parsep}{0pt}\setlength{\topsep}{0pt}
  \item Shot order follows a clear story or explanation.
  \item The same product and context are maintained throughout.
  \item Scene changes support a coherent narrative or demo.
\end{itemize}
&
\begin{itemize}\setlength{\itemsep}{0pt}\setlength{\parsep}{0pt}\setlength{\topsep}{0pt}
  \item Scenes feel random or out of order.
  \item Products or contexts change without explanation.
  \item Transitions break the story or confuse the viewer.
\end{itemize}
\\ \hline

\parbox[t]{0.17\textwidth}{\centering \textbf{Script--Visual}\\\textbf{Alignment}} &
\begin{itemize}\setlength{\itemsep}{0pt}\setlength{\parsep}{0pt}\setlength{\topsep}{0pt}
  \item The script accurately describes what appears on screen.
  \item Key selling points are shown at the right visual moments.
  \item No hallucinated objects, brands, or functions.
\end{itemize}
&
\begin{itemize}\setlength{\itemsep}{0pt}\setlength{\parsep}{0pt}\setlength{\topsep}{0pt}
  \item The script mentions things not visible in the video.
  \item Important visual content is not reflected in the script.
  \item Timing between narration and visuals is clearly off.
\end{itemize}
\\ \hline

\parbox[t]{0.17\textwidth}{\centering \textbf{BGM--Visual}\\\textbf{Compatibility}} &
\begin{itemize}\setlength{\itemsep}{0pt}\setlength{\parsep}{0pt}\setlength{\topsep}{0pt}
  \item Music tempo matches motion and editing rhythm.
  \item Overall mood fits the product and scenario.
  \item Volume and energy support, rather than overshadow, the content.
\end{itemize}
&
\begin{itemize}\setlength{\itemsep}{0pt}\setlength{\itemsep}{0pt}\setlength{\topsep}{0pt}
  \item Music is too fast/slow compared with visual rhythm.
  \item Style or emotion of the music feels inappropriate.
  \item Sudden changes in volume or style that distract the viewer.
\end{itemize}
\\ \hline

\parbox[t]{0.17\textwidth}{\centering \textbf{Overall}\\\textbf{Attractiveness}} &
\begin{itemize}\setlength{\itemsep}{0pt}\setlength{\parsep}{0pt}\setlength{\topsep}{0pt}
  \item The ad is engaging and easy to follow.
  \item Product value and advantages are clearly conveyed.
  \item You would be more willing to consider buying the product.
\end{itemize}
&
\begin{itemize}\setlength{\itemsep}{0pt}\setlength{\parsep}{0pt}\setlength{\topsep}{0pt}
  \item The ad feels dull, confusing, or unconvincing.
  \item Product value is unclear or poorly communicated.
  \item You would not be motivated to learn more or purchase.
\end{itemize}
\\ \hline

\end{tabular}
\end{table*}







